\documentclass[11pt,3p,review,authoryear]{elsarticle}
\usepackage[utf8]{inputenc} 
\usepackage[T1]{fontenc}    
\usepackage{hyperref}       
\usepackage{url}            
\usepackage{booktabs}
\usepackage{amsfonts}       
\usepackage{nicefrac}       
\usepackage{microtype}      
\usepackage{lipsum}
\usepackage{fancyhdr}
\usepackage{mdframed}
\usepackage{amsmath,amssymb,mathtools}
\usepackage{algorithm}
\usepackage{algpseudocode}
\usepackage{graphicx,float}
\usepackage{subfig}
\usepackage{bm} 
\usepackage{amsthm}
\usepackage{pifont}
\usepackage{xcolor}
\usepackage{bm}
\usepackage{bbm}
\newtheorem{theorem}{Theorem}
\newtheorem{lemma}{Lemma}
\newtheorem{definition}{Definition}

\hypersetup{
    colorlinks,
    linkcolor={red!50!black},
    citecolor={blue!50!black},
    urlcolor={blue!80!black}
}
\algrenewcommand\algorithmicrequire{\textbf{Input:}}
\algrenewcommand\algorithmicensure{\textbf{Output:}}

\journal{}
\begin{document}
\begin{frontmatter}

\title{A general framework for multi-step ahead adaptive conformal heteroscedastic time series forecasting}

\author[add1]{ Martim Sousa \corref{mycorrespondingauthor}}
\cortext[mycorrespondingauthor]{Corresponding author}
\ead{martimsousa@ua.pt}
\author[add1]{Ana Maria Tom{é}}
\ead{ana@ua.pt}
\author[add1]{Jos{é} Moreira}
\ead{jose.moreira@ua.pt}
\address[add1]{IEETA/DETI, University of Aveiro, Aveiro 3810-193, Portugal}

\begin{abstract}
This paper introduces a novel model-agnostic algorithm called adaptive ensemble batch multi-input multi-output conformalized quantile regression (\texttt{AEnbMIMOCQR}) that enables forecasters to generate multi-step ahead prediction intervals for a fixed pre-specified miscoverage rate $\alpha$ in a distribution-free manner. Our method is grounded on conformal prediction principles, however, it does not require data splitting and provides close to exact coverage even when the data is not exchangeable. Moreover, the resulting prediction intervals, besides being empirically valid along the forecast horizon, do not neglect heteroscedasticity. \texttt{AEnbMIMOCQR} is designed to be robust to distribution shifts, which means that its prediction intervals remain reliable over an unlimited period of time, without entailing retraining or imposing unrealistic strict assumptions on the data-generating process.
Through methodically experimentation, we demonstrate that our approach outperforms other competitive methods on both real-world and synthetic datasets. The code used in the experimental part and a tutorial on how to use \texttt{AEnbMIMOCQR} can be found at the following GitHub repository: \url{https://github.com/Quilograma/AEnbMIMOCQR}.
\end{abstract}

\end{frontmatter}

\section*{List of symbols}
\label{symbolstable}

\begin{table}[H]
\begin{tabular}{ll}
Notation & Description \\
\midrule
$\mathcal{A}$ & arbitrary regression algorithm (e.g., lightGBM) \\
$x$ &  a scalar \\
$\bm{x}$ &  a vector \\
$ \bm{X}$ & a tensor or random vector\\
$\alpha$ & miscoverage rate, between 0 and 1 \\
$\alpha_t$ & miscoverage rate, between 0 and 1 at the \textit{t}-th timestep. \\
$\bm{\epsilon}$ & set of \textit{non-conformity scores} \\
Quantile($\bm{\epsilon}$; $1-\alpha$) & $1-\alpha$ quantile of $\bm{\epsilon}$  \\
$C_{1-\alpha} (\bm{x_n})$ & oracle $1-\alpha$ prediction interval on input $\bm{x_n}$ \\
$\hat{C}_{1-\alpha}^{(t)} (\bm{x_{n}})$ & estimated $1-\alpha$ prediction interval on input $\bm{x_n}$ at the \textit{t}-th timestep\\
$Q_{\tau}(\bm{x})$ & population conditional quantile for $\tau \in (0,1)$ on input $\bm{x}$\\
$\hat{Q}_{\tau}(\bm{x})$ & conditional quantile estimate for $\tau \in (0,1)$ on input $\bm{x}$\\
$y_{t,h}$ & observation at the \textit{t}-th timestep for the  \textit{h}-th horizon\\
$X$ &  Random variable  \\
$\mathbbm{1}$ & Indicator function \\
\end{tabular}
\end{table}

\section{Introduction}
\label{sec1}
Recent advances in time series forecasting research have been driven by the increasing demand for higher forecast accuracy in complex settings, as set by modern big data applications \citep{79,80}. While considerable effort has been dedicated to identifying the most effective approaches, with the widely recognized M forecasting competition, initiated in 1982 \citep{82}, serving as a arena for these breakthroughs, quantifying the uncertainty of such predictions has received diminished attention. Indeed, the inclusion of prediction intervals (PIs) began to be considered in forecasting competitions only from the M4 competition \citep{81} onward. Another example of this discrepancy was observed in the M5 forecasting competitions. The M5 "Accuracy" competition \citep{83}, which centered on optimizing \textit{point predictions}, garnered significant interest and participation, with a staggering $7,092$ participants vying for top honors. In stark contrast, the M5 "Uncertainty" competition \citep{15}, which aimed to assess the quality of estimated conditional quantiles, drew considerably less attention, involving only $1,137$ participants despite offering the same prize incentives. 

Nevertheless, this seemingly diminished interest does not align with the actual relevance of probabilistic forecasting, which plays a pivotal role in numerous high-stakes domains, including finance \citep{87}, healthcare planning \citep{88,91}, energy \citep{90,89}, and urban planning \citep{92,93}, urging a greater focus on the topic. Probabilistic forecasting enables decision-makers to improve their decision-making by incorporating uncertainty, which is overlooked by \textit{point predictions}. A common form of probabilistic forecast that we will delve into is a PI, which defines a specific range within which the actual future value is anticipated to lie, with a predetermined level of probability (e.g., 0.9). It differs from a confidence interval \citep{74} in that it takes into consideration both \textit{aleatory} and \textit{epistemic} uncertainties \citep{1}.

While the advantages of PIs concerning informed decision-making are undeniable, their construction is anything but straightforward as time series are non i.i.d.. A recent paper \citep{86} alluded to the need for a distribution-free framework that generates PIs for time series data, along with provable guarantees for interval coverage and model-agnosticity properties so that it can be wrapped around complex and flexible models such as lightGBM \citep{84} that do not have an inherent method for generating PIs. In addition to the aforementioned criteria, a recent study \citep{94} draws out attention to the importance of PIs which handle heteroscedasticity.
Bellow, we outline all properties this framework should comply with: 

\begin{enumerate}
    \item Mitigating the reliance on strict distributional assumptions;
    \item Ensuring the validity of the generated PIs non-asymptotically;
     \item Handling heteroscedasticity;
    \item Ability to generate PIs for an arbitrary forecast horizon;
    \item Model-agnosticity;
    \item Minimizing the width of the generated PIs;
    \item Reliability under distribution shift without entailing retraining.
\end{enumerate}

In this paper, we embrace the challenge of presenting a framework designed to comprehend all of the aforementioned properties. Specifically, for an arbitrary forecast horizon $H$ we seek to ensure validity step-wise in the sense of (\ref{valditystepwise}), where $y_{t+h}$ is the unknown h-step ahead, $[\hat{y}_{t+h}^{L},\hat{y}_{t+h}^{U}]$ is the generated h-step ahead PI, and $\alpha \in (0,1)$ is the so-called miscoverage rate. 

\begin{equation}
\label{valditystepwise}
\mathbb{P}\{y_{t+h} \in [\hat{y}_{t+h}^{L},\hat{y}_{t+h}^{U}] \} = 1-\alpha , \quad \forall h \in\{1,2...,H\}   
\end{equation}

\subsection{Defining quality in prediction intervals}
\begin{figure}[H]
    \centering
    \begin{minipage}[t]{0.5\textwidth}
        \centering
        \includegraphics[width=\textwidth]{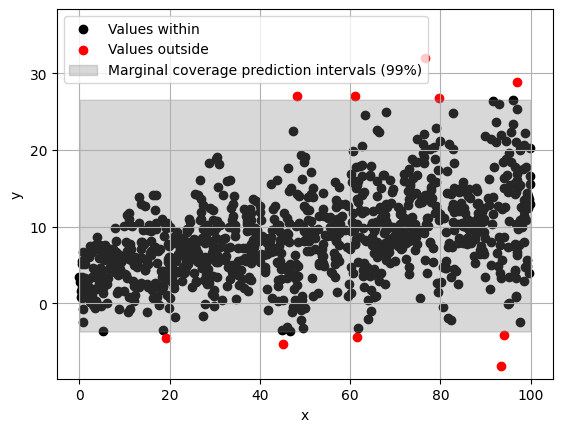}
        \caption{Marginally valid PIs for $\alpha=0.01$.}
        \label{MarginalCoveragePIs99}
    \end{minipage}%
    \begin{minipage}[t]{0.5\textwidth}
        \centering
        \includegraphics[width=\textwidth]{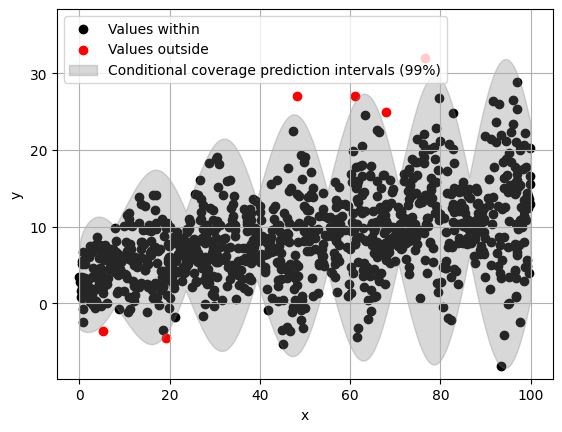}
        \caption{Conditionally valid PIs for $\alpha=0.01$.}
        \label{ConditionalCoveragePIs99}
    \end{minipage}
\end{figure}

\begin{figure}[H]
    \centering
    \begin{minipage}[t]{0.5\textwidth}
        \centering
        \includegraphics[width=\textwidth]{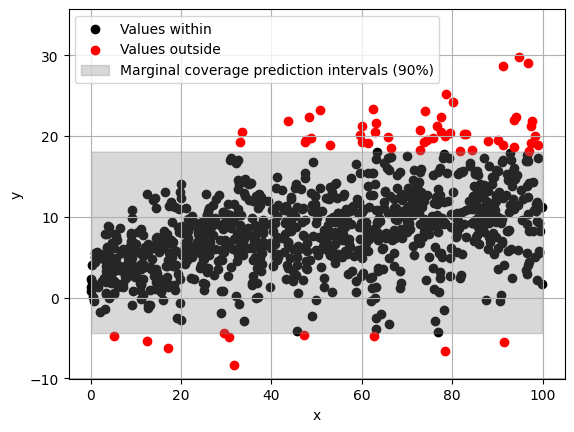}
        \caption{Marginally valid PIs for $\alpha=0.1$.}
        \label{MarginalCoveragePIs90}
    \end{minipage}%
    \begin{minipage}[t]{0.5\textwidth}
        \centering
        \includegraphics[width=\textwidth]{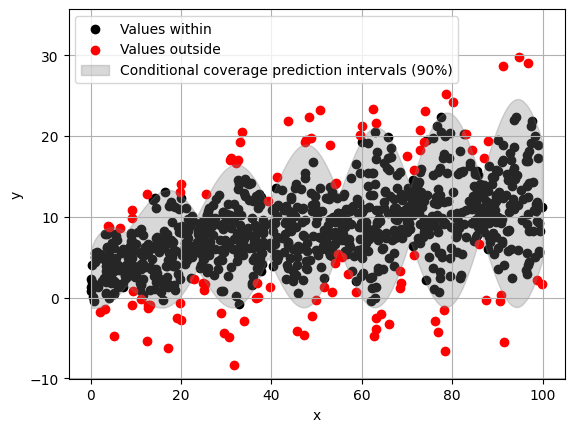}
        \caption{Conditionally valid PIs for $\alpha=0.1$.}
        \label{ConditionalCoveragePIs90}
    \end{minipage}
    \caption*{Data generated according to $y = 2+ \sqrt{x} + \epsilon(x)$, where $\epsilon(x)=\mathcal{N}(0,2+|0.6\sqrt{x}\cos{0,0.2x}|)$ is an heteroscedastic aleatory noise and $x \in[0,100]$. A total of 1000 observations are present in each plot.}
\end{figure}

We will now delve into establishing the key attributes that constitute a high-quality PI, not exclusively tied to time series forecasting, but more generally any regression task. To this end, consider an unseen pair of covariates and target, denoted as $(\bm{x_{n+1}}, y_{n+1})$; the foremost and arguably most crucial requirement that a high-quality PI should meet is validity, as outlined in Definition (\ref{marginalcoverage}). It refers to the alignment between the desired and observed nominal coverage. A valid, sometimes also referred to as calibrated PI, ensures the specified nominal coverage—neither more nor less. For instance, for a 90\% PI, around 90\% of real observations should fall within. Another requirement is sharpness. A sharp PI is as narrower as possible to be informative. Moreover, a frequently overlooked yet essential criterion that PIs should also comply with is the ability to handle heteroscedasticity. Heteroscedasticity occurs when the error variance exhibits variation across the covariates space. A clear understanding of the distinction between marginal and conditional coverage is vital in order to grasp why marginally valid PIs may not be sufficient. Essentially, conditionally valid PIs adapt to the varying error variance in the covariates space, as stipulated in Definition (\ref{conditionalcoverage}). To help comprehension of these concepts, two plots, Figures \ref{MarginalCoveragePIs99} and \ref{ConditionalCoveragePIs99}, are presented, pointing the differences between these two types of PIs. While in Figure \ref{ConditionalCoveragePIs99}, the PIs adapt to the local uncertainty of the given input, in Figure \ref{MarginalCoveragePIs99}, they fail to do so. Additionally, ensuring conditional coverage implies that miscoverage of the target variable is equally likely to occur across the covariates space. This becomes clearer if we increase $\alpha$ from $0.01$ to $0.1$, as shown in Figures \ref{MarginalCoveragePIs90} and \ref{ConditionalCoveragePIs90}, where we notice that in Figure \ref{MarginalCoveragePIs90} the ratio of miscoverage for $x \geq 40$ is much greater in comparison to $x<40$.

Thinking in a real-world setting where $x$ could represent a person's age, deploying PIs as depicted in Figure \ref{MarginalCoveragePIs90} would raise severe ethical concerns in practice, despite satisfying Definition (\ref{marginalcoverage}).

\begin{definition}[Marginal validity (also known as marginal coverage)]
\label{marginalcoverage}
A PI is termed valid if for an unseen pair of covariates and target $(\bm{x_{n+1}}, y_{n+1})$
\begin{equation}
    \mathbb{P}\left\{y_{n+1} \in \hat{C}_{1-\alpha}(\bm{x_{n+1}})\right\}= 1-\alpha
\end{equation}
\end{definition}

\begin{definition}[Conditional validity (also known as conditional coverage)]
\label{conditionalcoverage}
A PI is termed conditional valid if for an unseen pair of covariates and target $(\bm{x_{n+1}}, y_{n+1})$
\begin{equation}
    \mathbb{P}\left\{y_{n+1} \in \hat{C}_{1-\alpha}(\bm{x_{n+1}})| \bm{X_{n+1}} = \bm{x_{n+1}}\right\}  = 1- \alpha.
\end{equation}
\end{definition}

\subsection{Literature review}
Selecting the right approach to produce multi-step ahead PIs can be strenuous given the multitude of available approaches  \citep{47,23}. Certain traditional methods have an in-built theoretical way to do so \citep{2,17,3}. Nevertheless, their applicability is constrained by the reliance on robust parametric distributional assumptions. For example, PIs generated by  autoregressive integrated moving average (\texttt{ARIMA}) are based on the premises of linearity, independence of residuals, homoscedasticity, normality, and stationarity after differencing, dissuading numerous practitioners from applying these methods in practical, real-world scenarios, primarily because of the formidable task of confirming these assumptions, which are seldom fulfilled. Employing these methods without proper verification is even more detrimental, as it most likely results in unreliable PIs.

In the M5 UQ competition \citep{15}, one of the significant findings was that winning teams utilized simple machine learning (ML) models leveraging their inherent flexibility. Consequently, an advisable direction is to prioritize model-agnostic UQ methods, capable of encompassing a wide array of statistical models, particularly ML-based. These comprehend bootstrapping \citep{29,78}, deep ensembles \citep{48}, dropout \citep{49}, empirical \citep{4}, Bayesian \citep{77,35,46}, Gaussian processes \citep{50}, and quantile regression (QR) \citep{30,31}. However, it is crucial to note that, in isolation, these methods generally do not consistently achieve validity, as demonstrated in a comprehensive independent experiment \citep{41}. The authors of the aforementioned experiment, attribute this inconsistency between the desired and actual out-of-sample nominal coverage to the violation of certain assumptions that are inherent to some classes of methods. To overcome this limitation, the authors promote (CP) \citep{43,44,45} as a general calibration procedure for methods that deliver poor results without a calibration step. 

Unlike the aforementioned methods, CP ensures marginal coverage in finite samples in a distribution-free manner under the mild assumption of exchangeability, a slightly weaker assumption compared to i.i.d., that is already implicit while training most statistical models. Despite this advantage, PIs built uniquely from CP may have fixed or weakly varying width regardless of the error variance input-wise \citep{6,51,52}. To address this limitation, a method called conformalized quantile regression (CQR) \citep{14} was proposed. CQR employs CP along with QR to correct its predictive bands, inheriting the most attractive features of both, resulting in valid PIs that account for the present heteroscedasticity to a certain extend.               

In time series data, the assumption of exchangeability of CP may not hold, which can compromise the reliability of utilizing CP under such circumstances. Moreover, exchangeability implies identically distributed making it unsuitable in case of a distribution shift. Currently, CP has already been extended to handle time series \citep{12,7}. However, after carefully analyzing these methods, we identified opportunities for further improvement. Concretely, the proposed algorithm adapts faster to distribution shifts while delivering shorter PIs along the forecast horizon.               
\subsection{Contributions}
This paper addresses the criteria presented above by presenting an algorithm that constructs distribution-free PIs for volatile time series data. These PIs converge to valid over an unlimited period of time along the forecast horizon while considering heteroscedasticity. In essence, the key contributions of this paper are:

\begin{itemize}
    \item A novel algorithm, called \texttt{AEnbMIMOCQR}, which combines three existing methods: ensemble batch conformalized quantile regression (\texttt{EnbCQR}) \citep{7}, adaptive conformal inference (ACI) \citep{11}, and the multi-input multi-output (MIMO) forecasting strategy \citep{5}. Similarly to ensemble batch prediction intervals (\texttt{EnbPI}) \citep{12}, \texttt{AEnbMIMOCQR} does not require data splitting, having the advantage of handling heteroscedasticity by utilizing CQR \citep{14}.
    \texttt{AEnbMIMOCQR} has the ability of adapting to distribution shifts through post-training feedback received in batches. This is achieved by using a varying $\alpha_t$ parameter for the miscoverage rate and a sliding window of empirical residuals.  Consequently, \texttt{AEnbMIMOCQR} is capable of ensuring nearly valid PIs even on adverse settings. Finally, \texttt{AEnbMIMOCQR} utilizes the MIMO strategy to obtain $H$-step ahead PIs directly, which most likely has a positive effect on narrowing PIs width due to not accumulating errors unlike other forecasting strategies.
    \item Theoretically, \texttt{AEnbMIMOCQR} inherits the same asymptotic properties of ACI and \texttt{EnbPI}. Therefore, PIs generated by \texttt{AEnbMIMOCQR} are asymptotically valid under mild assumptions.
    \item Empirically, \texttt{AEnbMIMOCQR} nearly delivers the desired nominal coverage. Moreover, we evaluated the performance of \texttt{AEnbMIMOCQR} on both real-world and synthetic datasets and compared its PIs performance with suitable performance measures against those produced by state-of-the-art methods, namely \texttt{EnbPI} and \texttt{EnbCQR} as well as  \texttt{ARIMA} to serve as a baseline. Our experiments demonstrated that \texttt{AEnbMIMOCQR} adapts quicker to distribution shifts compared to \texttt{EnbPI} and \texttt{EnbCQR}, which is reflected in terms of a closer gap between observed and desired nominal coverage. Additionally, the width of the PIs generated by \texttt{AEnbMIMOCQR} are shorter and adapt to the local difficulty of the input, which suggests suitability for handling heteroscedasticity.
    \item Although this paper solely tackles univariate time series, \texttt{AEnbMIMOCQR} can be easily generalized to cope with multivariate time series and hints are provided to do so. Furthermore, \texttt{AEnbMIMOCQR} can be employed as a replacement for CQR in volatile regression settings, not necessarily time series, and for unsupervised anomaly detection tasks. 
    
\end{itemize}

\subsection{Paper outline}
The remainder of this paper is structured as follows: Section \ref{secback} provides a brief summary of the necessary background required to understand \texttt{AEnbMIMOCQR}, while Section \ref{secaenbmimocqr} introduces it in detail. In Section \ref{secexp}, we describe the datasets, software, parameters, and performance measures used to run the algorithms. Section \ref{secra} presents and analyzes the results obtained from our experiments. Finally, Section \ref{secdis} summarizes the main findings of the study and discusses possible future research directions.

The notation used throughout this paper can be consulted at the beggining of the paper.

All mathematical details used to prove the theorems have been omitted from the main text for brevity, but are available in \ref{proofs}. Additionally, pseudocode for the competitive algorithms can be found in \ref{algos}.

\section{Background}
\label{secback}
\subsection{Forecasting strategies}
When faced with a multi-step ahead forecasting problem, it is crucial to determine the most effective forecasting strategy \citep{5}. While this is a complex and continually evolving area of research that goes beyond the scope of this paper, we will briefly present and discuss two popular methods: recursive and MIMO. The former is used on \texttt{EnbPI} and \texttt{EnbCQR} while \texttt{AEnbMIMOCQR} uses the latter. 

\subsubsection{Recursive}
Suppose that from the following time series $\{y_1,y_2,...,y_n\}$ we want to estimate a function $f:\mathbb{R}^{p} \rightarrow \mathbb{R}$ to predict the next observation based on its $p$ \textit{lags}. How can this be achieved and how to use it to make $H$-step ahead \textit{point predictions} from an abstract regression algorithm $\mathcal{A}$?

First, we should convert the time series to a supervised learning problem as follows

\begin{equation}
\begin{bmatrix}
y_{1} & y_{2} & \cdots & y_{p} & \vert & y_{p+1} \\
y_{2} & y_{3} & \cdots & y_{p+1} & \vert & y_{p+2} \\
\vdots & \vdots & \ddots & \vdots & | & \vdots \\
y_{n-p} & y_{n-p+1} & \cdots & y_{n-1} & | & y_{n} \\
\end{bmatrix},
\end{equation}
where the left-hand side are covariates and the right-hand side is the target. This matrix can be constructed by stacking $B_i=(\underbrace{(y_i,y_{i+1},...,y_{i+p-1})}_{\text{Covariates}},\underbrace{y_{i+p}}_{\text{Target}}), \quad 1\leq i \leq n-p$ vertically.

Second, we estimate a regression function from the training data as $\hat{f} = \mathcal{A}(B_1,...,B_{n-p})$.

Finally, multi-step ahead \textit{point predictions} are computed from the following equations

\begin{equation}
\hat{y}_{n+h}=
\begin{cases}
\hat{f}(y_{n-p+1},...,y_n), \quad \mbox{for } h=1, \\
\hat{f}(y_{n-p+h},...,y_{n},\hat{y}_{n+1},...,\hat{y}_{n+h-1}), \quad \mbox{for } h\in\{2,...,p\},\\
\hat{f}(\hat{y}_{n+h-p},...,\hat{y}_{n+h-1}), \quad \mbox{for } h \in \{p+1,...,H\}.
\end{cases}
\label{recstrategy}
\end{equation}

The major flaw of this approach is that the estimation $\hat{f}$ was derived from a training set of actual \textit{lag} values. However, when the model is used to generated multi-step ahead \textit{point predictions} on new data in the out-of-sample phase, the prediction at each step is included as input for the next step, leading to accumulation of errors as the forecast horizon ($h$) increases. It is worth noting that for $h>p$, the input of the model only consists of forecasts, which can further exacerbate the issue.

\subsubsection{MIMO}
The MIMO strategy also involves converting time series data into a supervised learning problem, as shown in (\ref{mimostrategy}). However, unlike the recursive strategy, the target variable is a vector rather than a scalar, therefore $F:\mathbb{R}^{p} \rightarrow \mathbb{R}^{H}$ is multi-output. This has the advantage of avoiding error accumulation over a large forecast horizon. To make a forecast for the entire horizon, it only needs to incorporate the past $p$ observations to then forecast the whole forecast horizon in a single step.

Despite this appeal, the MIMO strategy has some bottlenecks. First, the structure of the model is more complex and therefore may require more training data. Second, the stochastic dependency between the observations of the forecast horizon is lost, which may lead to decreased forecast accuracy. Lastly, the regression algorithm $\mathcal{A}$ must be able to handle multi-output targets.     

\begin{equation}
\label{mimostrategy}
\begin{bmatrix}
y_{1} & y_{2} & \cdots & y_{p} & \vert & y_{p+1}& \cdots & y_{p+H}  \\
y_{2} & y_{3} & \cdots & y_{p+1} & \vert & y_{p+2}&  \cdots & y_{p+H+1} \\
\vdots & \vdots & \ddots & \vdots & | & \vdots & \cdots & \vdots \\
y_{n-H-p} & y_{n-H-p+1} & \cdots & y_{n-H-1} & \vert & y_{n-H}& \cdots & y_{n} \\
\end{bmatrix}
\end{equation}

Empirical studies have shown that multi-output strategies tend to perform better in multi-step ahead forecasting \citep{5,20}. However, this is not an universal rule, as confirmed in \citep{21}, following the "horses for courses" \citep{22} principle.

\subsection{Conformal prediction}
Currently, CP covers a broad range of ML problems, including regression, classification, unsupervised anomaly detection, and time series  \citep{45}. The strongest property of CP is its ability to ensure marginal coverage in finite samples while being model-agnostic and distribution-free. No assumptions are required beyond exchangeability. Several variants to employ CP comprehend transductive \citep{55}, jackknife+ \citep{62}, cross-conformal \citep{57} and inductive \citep{53}. Although some variants can potentially be more statistically efficient, this comes at the cost of a potentially more complex and computationally demanding algorithm. Therefore, for the sake of simplicity and space, only the inductive approach is introduced here. Henceforth referred to as CP. Following is the general outline of CP:

\begin{mdframed}
\begin{enumerate}
    \item Estimate a model $\hat{f}$ on a training dataset or use a pre-trained off-the-shelf model;
    \item Define a symmetric heuristic notion of uncertainty denoted as 
    $s: \mathcal{X} \times \mathcal{Y} \rightarrow \mathcal{A} \subseteq \mathbb{R}$ usually referred to as the \textit{non-conformity score function}, where larger scores encode worse agreement between pairs $(\bm{x},y)\in \mathcal{X} \times \mathcal{Y}$;
    \item Compute $\bm{\epsilon}=\{\epsilon_1,...,\epsilon_n\}=\{s(\bm{x_{1}},y_1),...,s(\bm{x_{n}},y_n)\}$ \textit{non-conformity scores} on a calibration dataset, not seen by the model during training, using the estimated model $\hat{f}$, and the \textit{non-conformity score function} $s$ applied on pairs of a calibration dataset $ \mathcal{D}_{\text{cal}}=\{(\bm{x_i},y_i)\}_{i=1}^{n}$;
    \item Compute $\hat{q}= \text{Quantile} \left( \bm{\epsilon};  \frac{\lceil (n+1)(1-\alpha)\rceil}{n}\right)$;
    \item For fresh covariates $\bm{x_{n+1}}$ deploy PIs as $\hat{C}_{1-\alpha}(\bm{x_{n+1}})=\{y \in \mathcal{Y}: s(\bm{x_{n+1}},y) \leq \hat{q}\}$.
\end{enumerate}
\end{mdframed}

CP is relatively straightforward as seen above. However, to account for the variability in the dataset, the calibration set should contain ideally a few thousand elements. Note that although the correct quantile in step 4, to account for a minor finite-sample correction, is $\frac{\lceil (n+1)(1-\alpha)\rceil}{n}$, for calibration's set sizes greater than one hundred, it is empirically equivalent to computing the $1-\alpha$ quantile. Hence, we will use the $1-\alpha$ quantile for simplicity's sake from now on.

Considering regression, one natural choice for the \textit{non-conformity score function} is the absolute error, which is defined as $\epsilon_i=|\hat{y}_i-y_i|$. Once we have computed the \textit{non-conformity scores} for all observations in the calibration set, we can compute their $1-\alpha$ quantile, denoted as $\hat{q}$. Afterwards, to obtain a PI for fresh covariates $\bm{x_{n+1}}$ with an unknown target value $y_{n+1}$, we first make a point prediction $\hat{f}(\bm{x_{n+1}})$. We then construct the PI by adding and subtracting the quantile $\hat{q}$ from the \textit{point prediction}, resulting in the interval $\hat{f}(\bm{x_{n+1}})\pm \hat{q}$. Based on Theorem (\ref{theoremmarginal}), which assumes exchangeability of the data-generating process, the provided PI is expected to achieve marginal validity. Additionally, the upper bound $(1-\alpha+\frac{1}{n+1})$ ensures that the PIs are not overly conservative, where n represents the calibration set size. As such, by increasing the size of the calibration set, CP converges to marginal validity.

In the upcoming section, we introduce a superior strategy for regression that accounts for heteroscedasticity. Notably, $\hat{f}(\bm{x_{n+1}})\pm \hat{q}$ remains constant in width, regardless of the input, as illustrated in Figure \ref{NaivePIs}.   

\begin{theorem}[Marginal coverage guarantee]
\label{theoremmarginal}
Let $(\bm{X_1},Y_1),...,(\bm{X_{n}},Y_{n})$ be exchangeable random vectors with no ties almost surely drawn from a distribution P, additionally if for a new pair $(\bm{X_{n+1}},Y_{n+1})$, $(\bm{X_1},Y_1),...,(\bm{X_{n+1}},Y_{n+1})$ are still exchangeable, then by constructing $C_{1-\alpha}(\bm{X_{n+1}})$ using CP, the following inequality holds for any \textit{non-conformity score function} $s:\mathcal{X} \times \mathcal{Y} \rightarrow \mathcal{A} \subseteq \mathbb{R}$ and any $\alpha \in (\frac{1}{n+1},1)$
\begin{equation}
    1-\alpha\leq \mathbb{P}\{Y_{n+1} \in C_{1-\alpha}(\bm{X_{n+1}})\} \le 1-\alpha+\frac{1}{n+1}.
\end{equation}
\end{theorem}

\subsection{Conformalized quantile regression}
\begin{figure}[H]
    \centering
    \begin{minipage}[t]{0.5\textwidth}
        \centering
        \includegraphics[width=\textwidth]{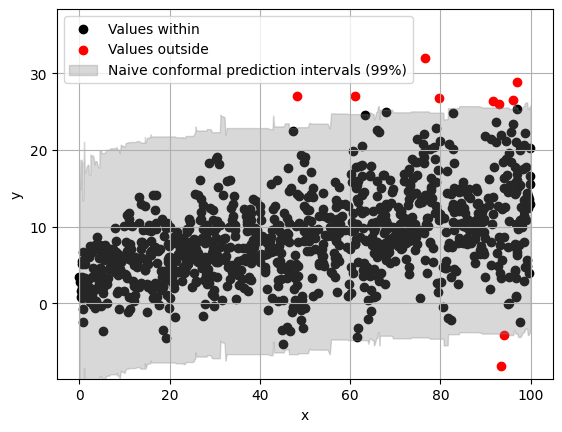}
        \caption{Naive conformal PIs for $\alpha=0.01$.}
        \label{NaivePIs}
    \end{minipage}%
    \begin{minipage}[t]{0.5\textwidth}
        \centering
        \includegraphics[width=\textwidth]{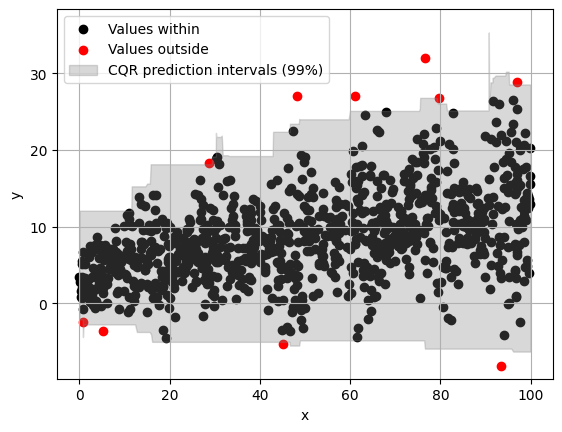}
        \caption{CQR PIs for $\alpha=0.01$.}
        \label{CQRPIs}
    \end{minipage}
    \caption*{Data generated according to $y = 2+ \sqrt{x} + \epsilon(x)$, where $\epsilon(x)=\mathcal{N}(0,2+|0.6\sqrt{x}\cos{0,0.2x}|)$ is an heteroscedastic aleatory noise and $x \in[0,100]$. A total of 1000 observations are present in each plot.}
\end{figure}

A better approach to build valid PIs is through the use of CQR \citep{14}. CQR utilizes QR \citep{30,31} to estimate conditional quantiles from the data, which are then corrected via CP to ensure marginal coverage. This approach allows for PIs of varying width depending on the input according to its local error variance, as depicted in Figure \ref{CQRPIs}.

Recall that the $\tau$-quantile of a conditional distribution $Y|\bm{X}$ is given by

\begin{equation}
    Q_{\tau}(\bm{x})=\inf{\{y \in \mathcal{Y}:F_{Y|\bm{X}}(y|\bm{x}) \ge \tau}\}.
\end{equation}

Based on this, the interval $[Q_{\alpha/2}(\bm{x}),Q_{1-\alpha/2}(\bm{x})]$ is conditionally valid given that 

\begin{equation}
\int_{Q_{\alpha/2}(\bm{x})}^{Q_{1-\alpha/2}(\bm{x})} f_{Y|\bm{X}}(y|\bm{x})\; dy=1-\alpha.    
\end{equation}

QR enables us to estimate these conditional population quantiles from the data by minimizing the so-called \textit{pinball loss} over the training set, which is mathematically expressed by
\begin{equation}
\label{pinball}
    \rho_{\tau}(y,\hat{f}(\bm{x}))=\max\left( \tau(y-\hat{f}(\bm{x})),  (\tau-1)(y-\hat{f}(\bm{x}))\right).
\end{equation}

To achieve the desired nominal coverage, two models have to be trained, each one set to minimize a \textit{pinball loss} function. One for $\tau=\alpha/2$ and another for $\tau=1-\alpha/2$. The resulting conditional quantile estimates are denoted as $\hat{Q}_{\alpha/2}(\bm{x})$ and $\hat{Q}_{1-\alpha/2}(\bm{x})$, respectively. 

Although the theory states that minimizing the \textit{pinball loss} is an asymptotic consistent estimator for the true conditional population quantile and therefore $\hat{Q}_{\tau}(\bm{x}) \overset{\text{P}}{\rightarrow} Q_{\tau}(\bm{x})$, this only holds under unverifiable regularity conditions such as the oracle conditional distribution being Lipschitz continuous \citep{32,34}. To further exacerbate the issue, extreme quantiles may be challenging to estimate due to data scarcity \citep{16}, resulting in poor quality estimates.

Given the information presented, it is highly likely that the interval $[\hat{Q}_{\alpha/2}(\bm{x}),\hat{Q}_{1-\alpha/2}(\bm{x})]$ will either deliver overcoverage or undercoverage. Although its unfeasible to correct these estimates in a way to match the oracle conditional quantiles, we can draw ideas from CP to turn these estimates into reliable valid PIs. To do so, \citet{14} devised a clever \textit{non-conformity score function} given by
\begin{equation}
\label{q_yhatCQR}
    s(\bm{x},y)=\max{\{\hat{Q}_{\alpha/2}(\bm{x})-y,y-\hat{Q}_{1-\alpha/2}(\bm{x})\}}.
\end{equation}

This \textit{non-conformity score function} assigns negative scores when the ground truth falls within the PI, and positive scores otherwise. The magnitude of the score is directly proporional to the distance from the ground truth to the nearest bound.
Through it, after computing all \textit{non-conformity scores} over the calibration dataset and computing $\hat{q}$, valid PIs are attained as 

\begin{equation}
    [\hat{Q}_{\alpha/2}(\bm{x})-\hat{q},\hat{Q}_{1-\alpha/2}(\bm{x})+\hat{q}].
\end{equation}

Note that in case of overcoverage $\hat{q}$ will be negative and positive in case of undercoverage. The role of $\hat{q}$ here is simply to make the smallest shift on the predictive bands so that the desired nominal coverage is ensured. This shift assumes homoscedasticity and can be improved \citep{85,75}. However, for the sake of simplicity, we will not delve into it further in this paper.

\subsection{Conformal prediction for time series}
The exchangeability assumption of CP, while mild, may not hold for time series data, since the order of observations does matter. To overcome this, several methods suggest rearranging time series in exchangeable blocks that preserve dependency \citep{58,59}. However, these blocks still have an inherent order and thus may not be fully exchangeable, hindering the efficacy of CP under these settings.

A more versatile method closely related to Jackknife+-after-bootstrap \citep{61} proposed in \citep{12} is \texttt{EnbPI}. \texttt{EnbPI} utilizes an \textit{ensemble learning} method called \textit{bagging} \citep{60} to mitigate \textit{overfitting} while generating in-sample \textit{non-conformity scores} on out-of-bag samples avoiding, therefore, data splitting. The set of \textit{non-conformity scores} is then updated in batches in the out-of-sample phase, making \texttt{EnbPI} adaptive to distribution shifts. Theoretically, the authors of \texttt{EnbPI} derived tight upper bounds for the absolute difference between the empirical and theoretical error distribution considering a stationary and strongly mixing error process. Empirically, their method ensured the desired coverage over time while \texttt{ARIMA} failed to do so.

Consider a training set $\mathcal{D}_{\text{train}}=\{(\bm{x_{i}},y_{i})\}_{i=1}^{n}$ and an arbitrary regression algorithm $\mathcal{A}$, in practice, \texttt{EnbPI} works as follows:

\begin{mdframed}
\begin{enumerate}
    \item Generate $B$ bootstrap sets from $\mathcal{D}_{\text{train}}$, referred to as $S_1,..,S_{B}$;
    \item Train the regression algorithm on the bootstrap sets to estimate predictors, $\hat{f}^{b} = \mathcal{A}(S_b), \quad \forall b \in\{1,..,B\}$;
    \item Loop through the training set and aggregate the predictors which did not use the \textit{i}-th row for training yielding \textit{point predictions} as $\hat{y}_i= \phi(\{\hat{f}^b(\bm{x_i}) \; | \; i \not\in S_b\})$, where $\phi$ is a suitable aggregator such as the mean;
    \item From the \textit{ensemble} \textit{point predictions}, build a set of \textit{non-conformity scores} ($\bm{\epsilon}$) from absolute errors $\epsilon_i=|\hat{y}_i-y_i|$;
    \item Compute $\hat{q} = \text{Quantile}(\bm{\epsilon}; 1-\alpha)$;
    \item For each new out-of-sample set of covariates $\bm{\hat{x}_{n+1}},...,\bm{\hat{x}_{n+H}}$, construct  \textit{ensemble point predictions} for each element of the forecast horizon as $\hat{y}_{n+h}= \phi(\{\hat{f}^b(\bm{\hat{x}_{n+h}})\}_{b=1}^B)$ to then output PIs as $\hat{y}_{n+h} \pm \hat{q}$. After delivering $H$-step ahead PIs and receiving external feedback about the ground truth values $y_{n+1},...,y_{n+H}$, store out-of-sample residuals as $\epsilon_{n+h}=|\hat{y}_{n+h}-y_{n+h}|$ in memory and update $\bm{\epsilon}$ by incorporating these \textit{non-conformity scores} while discarding the \textit{H}-th most oldest. Compute $\hat{q} = \text{Quantile}(\bm{\epsilon}; 1-\alpha)$ again. Repeat this step indefinitely after each $H$-step ahead PIs are produced.    
\end{enumerate}
\end{mdframed}

The strength of \texttt{EnbPI} lies in its adaptability, achieved through step 6, where the \textit{non-conformity set} is transformed into a sliding window of empirical residuals. However, it has a significant drawback when applied to univariate time series forecasting. This issue arises because the in-sample \textit{non-conformity scores} are computed using actual lag values to generate \textit{point predictions} whereas, during the out-of-sample phase, each set of covariates $\hat{\bm{x}}_{\bm{n+h}}$ is computed using the recursive strategy formula (as described in (\ref{recstrategy})). Due to this recursive nature, these values accumulate errors over time, in contrast to the in-sample \textit{non-conformity scores} which do not reflect this behaviour. Additionally, given that the size of the \textit{non-conformity set} $(\approx n)$ is usually much larger than the forecast horizon $(H)$, it may take a considerable amount of time before all in-sample \textit{non-conformity scores} can be discarded. This may result in significant differences between the desired nominal coverage and the observed coverage through a long period a time, despite converging to the specified coverage.

Another drawback of the \texttt{EnbPI} method is its constant PIs width within each batch. However, by utilizing the CQR approach, it is feasible to overcome this limitation. This results in the well-known \texttt{EnbCQR} \citep{7}.

\subsection{Adaptive conformal inference}
ACI \citep{11} is a method designed to extend CP to volatile non-exchangeable settings. The approach involves initializing $\alpha_1$ as $\alpha$ and sequentially updating a time-varying $\alpha_t$ on received feedback via the following equation

\begin{equation}
\alpha_{t+1}=\alpha_{t}+\gamma(\alpha-\mathbbm{1}\{y_t \not \in \hat{C}^{(t)}_{1-\alpha_t}(\bm{x_t})\}),
\end{equation}

where $\gamma$ is a learning rate that controls the method's adaptation speed to distribution shifts. Note that if the true value is outside the PI, we update the PI width by increasing $\alpha$ using the recursion $\alpha_{t+1}=\alpha_t+\gamma(\alpha-1)$. This results in a wider PI for the next iteration, as $\alpha_{t+1} < \alpha_{t}$. On the other hand, if the true value is within the PI, we reduce the width of the PI by increasing $\alpha$, since $\alpha_{t+1}=\alpha_t+\gamma\alpha > \alpha_t$.

Theoretically, the authors of ACI have demonstrated that for a non-decreasing quantile function, the following inequality holds

\begin{equation}
\left|\frac{1}{n}\sum_{t=1}^n \mathbbm{1}\{y_t \not \in \hat{C}^{(t)}_{1-\alpha_t}(\bm{x_t})\}-\alpha\right| \leq \frac{\max\{\alpha_1,1-\alpha_1\}+\gamma}{n\gamma},
\end{equation}

which implies that $\lim_{n\rightarrow\infty}\frac{1}{n}\sum_{t=1}^{n} \mathbbm{1}\{y_t \not \in \hat{C}^{(t)}_{1-\alpha_t}(\bm{x_t})\}=\alpha$, demonstrating the method's asymptotic consistency.

However, in practice, ACI has several bottlenecks. First, nothing prevents $\alpha_t <0$, which leads to an infinite PI, despite being highly unlikely to occur in practice, especially if $\gamma$ is small. Second, the learning rate $\gamma$, which is passed as a hyperparameter has strong impact on the method's efficiency. Lastly, given that the \textit{non-conformity set} remains unchanged, updating the miscoverage rate $\alpha$ alone may not be sufficient to handle substantial distribution shifts. 

Taking into account the last bottleneck of ACI, we have decided to integrate it with a sliding window of empirical residuals in our proposal to further improve adaptability.

\section{AEnbMIMOCQR}
\label{secaenbmimocqr}
We now introduce \texttt{AEnbMIMOCQR}, a method that closely follows the structure of \texttt{EnbPI} but offers several additional advantages. Firstly, \texttt{AEnbMIMOCQR} employs the MIMO strategy that always uses actual values as input, avoiding any mismatch between in-sample and out-of-sample \textit{non-conformity scores}. Secondly, it handles heteroscedasticity by targeting conditional quantiles via a multi-output version of CQR. Thirdly, for situations where $n>>H$ an optional sampling of the \textit{non-conformity set} without replacement before entering the out-of-sample phase is provided in order to decrease its size and thus improve adaptability. Finally, \texttt{AEnbMIMOCQR} leverages ACI along with a sliding window of \textit{non-conformity scores}, allowing it to quickly adapt to distribution shifts.

The complete pseudocode of \texttt{AEnbMIMOCQR} is presented in Algorithm (\ref{AEnbMIMOCQRalg}). The first part of the algorithm focuses on the in-sample phase (lines 1-12), while the second part (lines 17-26) is dedicated to the out-of-sample phase. We also assume that $n_{\text{test}} \equiv 0 \mod H$ for simplicity's sake. To simplify the notation, we have used pairs of vectors $(\bm{x_i},\bm{y_i})$, where $\bm{x_i}$ is the left-hand side and $\bm{y_i}$ the right-hand side of line $i$ as in (\ref{mimostrategy}). 

In terms of coverage, \texttt{AEnbMIMOCQR} PIs are asymptotic valid along the forecast horizon, in the sense of  (\ref{AEnbMIMOCQRass}).

\begin{align}
\label{AEnbMIMOCQRass}
    \lim_{n \rightarrow \infty }\frac{1}{n}\sum_{t=1}^n \mathbbm{1}\{y_{t,h} \in [\hat{y}_{t,h}^{L},\hat{y}_{t,h}^{U}]\}= 1-\alpha, \quad \forall h \in\{1,2...,H\}
\end{align}

Specifically, in the in-sample phase, out-of-bag \textit{non-conformity scores} are computed for each step ahead from the ensemble multi-output QR models following the \textit{non-conformity score function} presented in (\ref{q_yhatCQR}) as

\begin{equation}
    \epsilon_{t,h}^{\phi}=\max\{\hat{y}_{t,h}^{L}-y_{t,h},y_{t,h}-\hat{y}_{t,h}^{U}\}, \quad \forall h \in\{1,...,H\}.
\end{equation}

Afterwards, we provide multi-step ahead PIs as
\begin{equation}
    [\hat{y}_{t,h}^{L}-\hat{q}^{(h)},\hat{y}_{t,h}^{U}+\hat{q}^{(h)}], \quad \forall h \in\{1,...,H\},
\end{equation}
where $\hat{q}^{(h)}$ is the h-step ahead QR's CP correction.

Similarly to \texttt{EnbPI}, in the out-of-sample phase, the \textit{non-conformity set} is updated with new \textit{non-conformity scores} after generating $H$-step ahead PIs. In addition, ACI is utilized initializing $\alpha_h$ as $\alpha$, where $\alpha_h$ is the miscoverage rate for the $h$-step ahead. In line 16 of Algorithm (\ref{AEnbMIMOCQRalg}), the symbol $\bm{\epsilon_h}^{'}$ refers to the \textit{non-conformity set} obtained from $\bm{\epsilon_h}$ after sampling $T$ scores without replacement. The user specifies the dimension of $\bm{\epsilon_h}^{'}$ as $T$, which is chosen to be much smaller than the original dimension $(\approx n)$ of $\bm{\epsilon_h}$. This reduction in dimension is intended to improve the adaptability of the algorithm.

For instance, if $\bm{\epsilon_h}$ has a dimension of $10,000$ and $H=10$ for each $h$-step ahead, the update of the empirical quantiles after producing the first $H$-step ahead PIs and updating the \textit{non-conformity set} will be negligible. This is because the dimension of the \textit{non-conformity set} is much larger than $H$. However, by specifying $T=100$ and using the sampled version $\bm{\epsilon_h}^{'}$ instead of $\bm{\epsilon_h}$, the effect on the empirical quantiles will be more significant since the dimension is now much smaller and closer to $H$. Undoubtedly, information loss is inevitable in this process, however it is worth employing in practice considering the gains in adaptability.       

Overall, \texttt{AEnbMIMOCQR} is expected to comply with the criteria initially presented in Section \ref{sec1}, implying it should empirically deliver the desired nominal coverage along the forecast horizon even on the most adverse volatile settings.

\begin{algorithm}[H]
\caption{\texttt{AEnbMIMOCQR} algorithm}.
\label{AEnbMIMOCQRalg}
\begin{algorithmic}[1]
\Require A training set $\{(\bm{x_i},\bm{y_i})\}_{i=1}^{n}$, miscoverage rate $\alpha$, a multi-output QR algorithm $\mathcal{A}_{\tau}$, an aggregation function $\phi$, number of bootstrap models $B$, the forecast horizon $H$, $T$ observations to sample without replacement, a test set $\{(\bm{x_t}, \bm{y_t})\}_{t=n+1}^{n+n_{\text{test}}}$ with $\bm{y_{t}}$ revealed only at timestep $t+H$.
\Ensure PIs for $\bm{x_{n+1}}, \bm{x_{n+1+H}},..., \bm{x_{n+n_\text{test}}}$.
\For{$b\gets1,...,B$}
\State Sample an index set $S_b=(i_1,...,i_T)$ from indices (1,...,$n$)
\State Train $\hat{F}_{\alpha/2}^{b} \gets \mathcal{A}_{\alpha/2}(\{(\bm{x_i},\bm{y_i})\;|\;i \in S_b\})$
\State Train $\hat{F}_{1-\alpha/2}^{b} \gets \mathcal{A}_{1-\alpha/2}(\{(\bm{x_i},\bm{y_i})\;|\;i \in S_b\})$
\EndFor
\State $\bm{\epsilon}_h \gets \{\}, \quad \forall h \in\{1,...,H\}$

\For{$i \gets 1,...,n$}
\State $[\hat{y}_{i,1}^{L},...,\hat{y}_{i,H}^{L}] \gets \phi(\{(\hat{F}_{\alpha/2}^{b}(\bm{x_i})|\;i \not\in S_b\})$
\State $[\hat{y}_{i,1}^{U},...,\hat{y}_{i,H}^{U}] \gets \phi(\{(\hat{F}_{1-\alpha/2}^{b}(\bm{x_i})|\;i \not\in S_b\})$
\State $ \epsilon_{i,h}^{\phi} \gets \max{\{\hat{y}_{i,h}^{L}-y_{i,h},y_{i,h}-\hat{y}_{i,h}^{U}\}}, \quad \forall h \in\{1,...,H\}$
\State $\bm{\epsilon}_h \gets \bm{\epsilon}_h \cup \{\epsilon_{i,h}^{\phi}\}, \quad \forall h \in\{1,...,H\}$
\EndFor
\State $\gamma \gets 1/\max\{T,|\bm{\epsilon_1}|\}$
\State $\alpha_h \gets \alpha, \quad \forall h \in\{1,...,H\}$
\State $\hat{q}^{(h)} \gets \text{Quantile}(\bm{\epsilon_h};1-\alpha_h), \quad \forall h \in\{1,...,H\}$
\State  $\bm{\epsilon_h}^{'} \gets  \text{Sample} (T,\bm{\epsilon_h}), \quad \forall h \in\{1,...,H\}$
\For{$t\gets n+1,n+1+H,...,n+n_{\text{test}}$}
\State $[\hat{y}_{t,1}^{L},...,\hat{y}_{t,H}^{L}] \gets \phi(\{(\hat{F}_{\alpha/2}^{b}(\bm{x_t})\}_{b=1}^B)$
\State $[\hat{y}_{t,1}^{U},...,\hat{y}_{t,H}^{U}] \gets \phi(\{(\hat{F}_{1-\alpha/2}^{b}(\bm{x_t})\}_{b=1}^B)$
\State Return $\hat{C}_{1-\alpha_h}^{(t)}(\bm{x_t}) \gets [\hat{y}_{t,h}^{L}-\hat{q}^{(h)},\hat{y}_{t,h}^{U}+\hat{q}^{(h)}], \quad \forall h \in\{1,...,H\}$
\State $\epsilon_h^{(*)} \gets  \max{\{(\hat{y}_{t,h}^{L}-\hat{q}^{(h)})-y_{t,h},y_{t,h}-(\hat{y}_{t,h}^{U}+\hat{q}^{(h)})\}}, \quad \forall h \in\{1,...,H\}$
\State $\bm{\epsilon}_h^{'} \gets (\bm{\epsilon}_h^{'}-\{\epsilon_{1,h}\}) \cup \{\epsilon_{h}^{(*)}\}, \quad \forall h \in\{1,...,H\}$ and reset index of $\bm{\epsilon}_{h}$
\State    $\alpha_{h} \gets\alpha_{h}+\gamma(\alpha-\mathbbm{1}\{y_{t,h} \not\in \hat{C}_{1-\alpha}^{(t)}(\bm{x_t})\}), \quad \forall h \in\{1,...,H\}$
\State $\alpha_h\gets\max\{0,\min\{\alpha_h,1\}\}, \quad \forall h \in\{1,...,H\}$
\State Update $\hat{q}^{(h)} \gets \text{Quantile}(\bm{\epsilon_h^{'}};1-\alpha_h), \quad \forall h \in\{1,...,H\}$
\EndFor
\end{algorithmic}
\end{algorithm}

\subsection{AEnbMIMOCQR for multivariate time series}
Thus far, this article has solely tackled conformal univariate time series forecasting; however, in many applications, the covariates are lagged multidimensional features that intend to estimate a $H$-step ahead multidimensional target, therefore the traning set $\mathcal{D}_\text{train}=\{(\bm{X_i},\bm{Y_{i}})\}_{i=1}^{n}$ is used to estimate the desired function $F:\mathbb{R}^{p \times F_1} \rightarrow \mathbb{R} ^{H \times F_2}$, where $F_1 \geq F_2$ are the number of input and output features, respectively. \texttt{AEnbMIMOCQR} can handle this setting by training a suitable ML model (e.g. \citep{63}) set to minimize the \textit{pinball loss} and by also modifying Algorithm (\ref{AEnbMIMOCQRalg}) accordingly, so that we have $H \times F_2$ \textit{non-conformity sets}, $\alpha$s, and $\hat{q}$s, one per each feature and horizon.

In the interest of space, Algorithm (\ref{AEnbMIMOCQRalg}) will not be reproduced to contemplate this case since the generalization is straightforward from here. 
\section{Experimental design}
\label{secexp}
The goal of the experimental part is to compare the PIs generated by the \texttt{AEnbMIMOCQR} method against those generated by four other methods: \texttt{EnbPI}, \texttt{EnbCQR}, \texttt{ARIMA}, and \texttt{MIMOCQR}, which is a simplified version of \texttt{AEnbMIMOCQR} that does not adapt to distribution shifts and does not use \textit{bagging} \citep{60}. Although the quality of the PIs generated by the conformal methods depends on the underlying regression algorithm, finding the best architecture is outside the scope of this study. Instead, the same architecture is employed for all methods, enabling a fair comparison of their relative performance. Table (\ref{tablepre}) displays the global parameters used in the experiment. 

\begin{table}[H]
    \centering
    \begin{tabular}{cc}
         \hline
         \textbf{Parameter} & \textbf{value}  \\
         \hline
         $\alpha$ & 0.1\\
         Number of lags (p) & 40\\
         Forecast horizon (H) & 30 \\
         \hline
    \end{tabular}
    \caption{Global parameters used throughout the experiment.}
    \label{tablepre}
\end{table}

\subsection{Datasets}
To evaluate the efficacy of the methods, two experiments were conducted. The first experiment employed a real-world dataset, while the second experiment was conducted using a synthetic dataset to provide a more controlled setting since the aleatory noise distribution is known.
\subsubsection{Real-world}
The real-world dataset is the NN5 forecasting competition dataset \citep{NN5Dataset}. This dataset comprises 111 time series, with each time series consisting of 791 daily cash machine withdrawals recorded at a specific location in England. This dataset is a suitable choice for this analysis since, as shown in Table (\ref{adftesttable}), the Augmented Dickey-Fuller test \citep{64,65} indicates that a reasonable amount of time series are non-stationary for a maxlag of 40.

\begin{table}[H]
    \centering
    \begin{tabular}{cc}
         \toprule
         Significance level & \#series  \\
         \midrule
         0.1 & 92\\
         0.05 & 84\\
         0.01 & 67 \\
         \bottomrule
    \end{tabular}
    \caption{Number of stationary time series for different significance levels of the Augmented Dickey-Fuller test.}
    \label{adftesttable}
\end{table}
\subsubsection{Synthetic}
In order to create a controlled experimental environment, data is generated using (\ref{syntheticprocess}), where $\{Y_i\}_{i=1}^{40} \sim \text{U}(0,1)$ and $c_t = 0.1 + t/1000$. This allows for a rigorous comparison between methods, as we have direct access to the oracle PIs. Additionally, since the data-generating process is dynamic it enables us to assess how different methods adapt to distribution shifts.

\begin{equation}
\label{syntheticprocess}
    Y_t = \mathcal{N}\left(\log\left(\sum_{i=1}^{40}Y_{t-i}^2\right),c_t\log\left(\sum_{i=1}^{40}Y_{t-i}^2 \right)\right), \quad t=41,...,1041.
\end{equation}
\subsection{Experimental setup}
In this experiment, a multilayer perceptron \citep{67} regression algorithm with two hidden layers and 64 neurons in each layer was utilized. The models were estimated using ADAM \citep{66} as a optimizer for 1000 epochs. The architecture of the regression algorithm differed for \texttt{AEnbMIMOCQR} and \texttt{MIMOCQR}, both having an $H$-dimensional output layer, while \texttt{EnbCQR} and \texttt{EnbPI}, had a $1$-dimensional output layer. Additionally, while \texttt{EnbPI} was set to minimize the mean squared error, the other competitive methods were set to minimize the \textit{pinball loss} at the quantiles of interest. The mean function was used to aggregate 10 bootstrap models ($B=10 \text{ and } \phi=\text{'mean'}$) for all models. $T$ was set to 100 on \texttt{AEnbMIMOCQR} for both datasets. The number of testing observations was set to 390, therefore $n_{\text{test}}=390$, leaving the remainder for training. The experiment was conducted using the Python programming language along with the \texttt{tensorflow} \citep{68} and \texttt{pmdarima} \citep{69} libraries.

\subsection{Evaluation performance measures}
The quality of the delivered PIs in the real-world dataset was evaluated using two performance measures: prediction interval coverage probability (PICP), which assesses the calibration, and prediction interval normalized average width (PINAW), which measures the sharpness. The PICP performance measure can also be found in \citep{70,7} and is defined in (\ref{picp}), while the PINAW performance measure can be found in \citep{72,7} and is defined in (\ref{pinaw}). Note that different performance measure could be used such as the \textit{scalled pinball loss} \citep{15} or Winkler score \citep{76}. However, we prefer to report sharpness and calibration separately since for critical applications, ensuring the desired nominal coverage is the primary concern over shorter PIs. Moreover, performance measures such as the Winkler score are not adequate for evaluating PIs that aim to capture heteroscedasticity noises.        

In the synthetic dataset, a third performance measure inspired by insersection over union (IOU) was employed. This performance measure penalizes PIs that deviate from the oracle PIs, whether they are overly conservative or under conservative. We name it mean intersection over union (MIOU) since it is the average IOU over the test set. The formula is presented in (\ref{MIOU}). 

\begin{align}
    & \text{PICP} = \frac{1}{n_{\text{test}}} \sum_{t=1}^{n_{\text{test}}} \mathbbm{1}\left\{y_t \in [\hat{y}^{L}_t, \hat{y}^{U}_t]\right\}, \label{picp} \\
    & \text{PINAW} = \frac{1}{n_{\text{test}} (y_{\text{max}}-y_{\text{min}})} \sum_{t=1}^{n_{\text{test}}} (\hat{y}^{U}_t -\hat{y}^{L}_t),  \label{pinaw}\\
    & \text{MIOU} = \frac{1}{n_{\text{test}}} \sum_{t=1}^{n_{\text{test}}} \frac{|[\hat{y}^{L}_t, \hat{y}^{U}_t] \cap [y^{L}_t, y^{U}_t]|}{|[\hat{y}^{L}_t, \hat{y}^{U}_t] \cup [y^{L}_t, y^{U}_t]|}  = \frac{1}{n_{\text{test}}} \sum_{t=1}^{n_{\text{test}}} \frac{|[\max\{\hat{y}^{L}_t,y^{L}_t\}, \min\{\hat{y}^{U}_t,y^{U}_t\}]|}{|[\min\{\hat{y}^{L}_t,y^{L}_t\}, \max\{\hat{y}^{U}_t,y^{U}_t]|} \label{MIOU}.
\end{align}

\section{Results and analysis}
\label{secra}

\subsection{Real-world dataset}
\begin{figure}[H]
    \centering
    \includegraphics[scale=0.4]{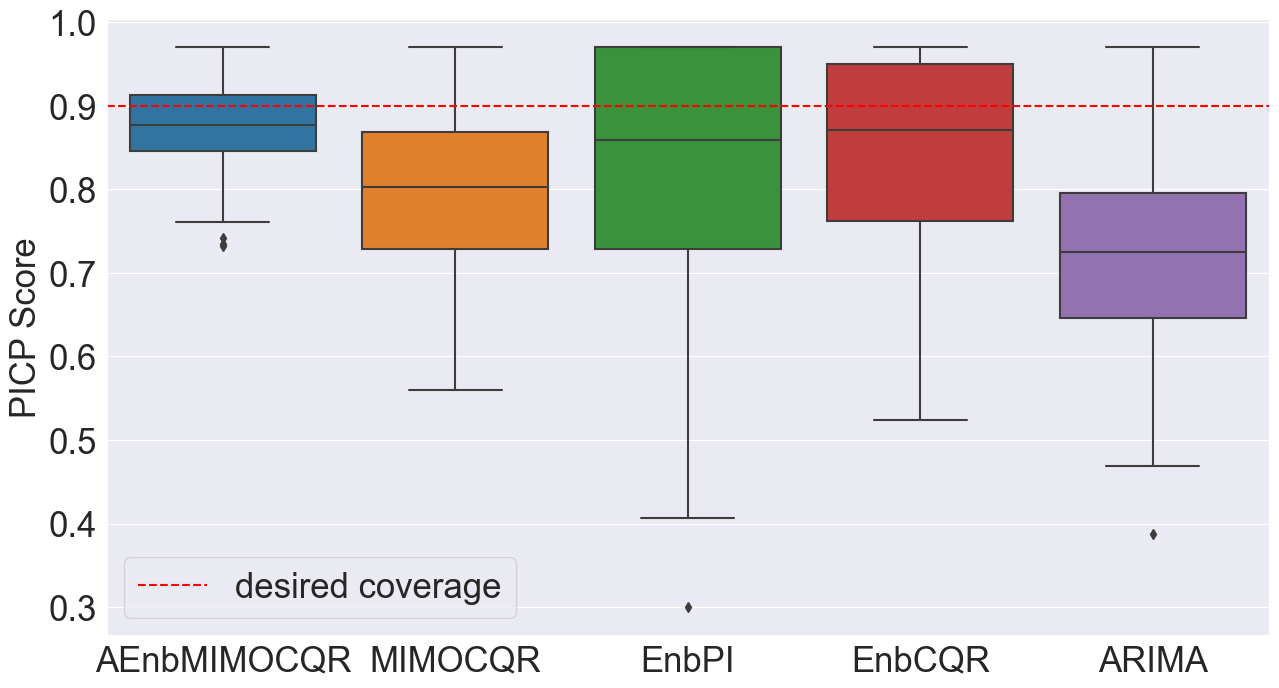}
    \caption{PICP score boxplot of all methods computed on the NN5 daataset.}
    \label{boxplotpicp}
\end{figure}

The results of all methods with the parameters specified in the last section are summarized in Table (\ref{resultstable1}). Figure \ref{boxplotpicp} shows the PICP boxplot scores for all methods. Since we are comparing 111 series and the chosen performance measures are scale-independent, we have calculated the average values across all series, according to (\ref{picpast}) and (\ref{pinawast}).

Upon inspecting Table (\ref{resultstable1}) results, it becomes evident that \texttt{AEnbMIMOCQR} outperforms its competitors in terms of observed coverage, as indicated by its $\text{PICP}^{*}$ score, which is closer to the desired nominal coverage of 0.9. Furthermore, its PIs width is lower than that of \texttt{EnbPI} and \texttt{EnbCQR}, most likely due to not accumulating errors over the forecast horizon. \texttt{ARIMA} method appears to be the furthest from achieving the desired nominal coverage. Although it produces wide PIs according to its $\text{PINAW}^{*}$, they seem to be mislocated. Given that the data-generating process is unknown, we cannot pinpoint the specific reasons for this issue. However, it is likely due to one or more violations of the many rigid assumptions underlying the method. As for the adaptive components that integrate \texttt{AEnbMIMOCQR} such as ACI and a sliding window of empirical residuals, they play a crucial role in achieving a 0.078 higher $\text{PICP}^{*}$ score by comparing it to \texttt{MIMOCQR}.

Moreover, Figure \ref{boxplotpicp} reveals that the spread of \texttt{AEnbMIMOCQR} is lower, which is expected based on the algorithm formulation. While \texttt{AEnbMIMOCQR} seeks to ensure exact nominal coverage and updates the $\alpha$ when the algorithm is either overcovering or undercovering, its competitors do not, justifying the shorter spread.

\begin{align}
    & \text{PICP}^{*}=\frac{1}{111} \sum_{i=1}^{111} \text{PICP}_i \label{picpast}\\
    & \text{PINAW}^{*}=\frac{1}{111} \sum_{i=1}^{111} \text{PINAW}_i \label{pinawast}
\end{align}

\begin{table}[H]
    \centering
\begin{tabular}{cccc}
\toprule
\textbf{Method} &  $\textbf{PICP}^{*}$ &      $\textbf{PINAW}^{*}$ \\
\midrule
\texttt{AEnbMIMOCQR} &  0.882 &  0.357  \\
\texttt{MIMOCQR} &  0.804 &  0.283  \\
\texttt{EnbPI} &  0.862 &  0.493  \\
\texttt{EnbCQR} &  0.866 &  0.370  \\
\texttt{ARIMA} &  0.723 &  0.403 \\
\bottomrule
\end{tabular}
\caption{Relative performance measures obtained on the real-world dataset.}
\label{resultstable1}
\end{table}

\subsection{Synthetic dataset}
The experiments conducted on the synthetic dataset have shown results that are consistent with those obtained on the real-world dataset. However, this additional experiment is key since relying solely on measures such as PICP and PINAW may not be sufficient if the goal is to assess conditional nominal coverage. In such cases, it is essential to consider other performance measures, such as MIOU, which can effectively take into account the heteroscedasticity of the dataset. Therefore, based on the evidence shown in Table (\ref{resultstable2}), \texttt{AEnbMIMOCQR} gets closer in achieving conditional nominal coverage, where a MIOU score of 1 signifies perfect conditional nominal coverage.    

\begin{table}[H]
    \centering
\begin{tabular}{cccc}
\toprule
\textbf{Method} &  \textbf{MIOU} \\
\midrule
\texttt{AEnbMIMOCQR} &  0.843  \\
\texttt{MIMOCQR} &  0.784   \\
\texttt{EnbPI} &  0.752   \\
\texttt{EnbCQR} &  0.796   \\
\texttt{ARIMA} &  0.622  \\
\bottomrule
\end{tabular}
\caption{MIOU scores on the synthetic dataset.}
\label{resultstable2}
\end{table}

\section{Discussion and conclusions}
\label{secdis}
In this paper, we proposed a novel algorithm called \texttt{AEnbMIMOCQR} that extends CP for multi-step ahead regression tasks without relying on exchangeability assumptions. Although previous work, such as \texttt{EnbPI} and \texttt{EnbCQR}, has explored the extension of CP to non-exchangeable data settings, our algorithm demonstrated faster adaptation to distribution shifts. Specifically, we observed more consistent observed coverage levels when compared to competing methods, as well as shorter and input-varying PIs. Moreover, using ACI, \textit{bagging}, and a sliding window of empirical \textit{non-conformity scores} revealed key as, contrarily to the simplified version, \texttt{MIMOCQR}, the observed coverage closely aligns with the desired nominal coverage. These results highlight the potential benefits of using \texttt{AEnbMIMOCQR} in real-world applications where distribution shifts and heteroscedasticity may occur, enhancing decision-making in critical applications.

Regardless of whether the data is heteroscedastic or not, we concur with the recommendation of \citet{7} to always use heteroscedastic methods such as \texttt{EnbCQR} and \texttt{AEnbMIMOCQR}. Even in cases where the data is homoscedastic, the PIs produced by \texttt{EnbCQR} will be similar to those of \texttt{EnbPI}. 
 
Topics worth exploring in the future are: (i) devise a clever mechanism on the choice of $\gamma$ in ACI, as it an important impact on the method's adaptation speed; (ii) comparing our approach against different forecasting strategies for multi-step ahead forecasting presented in the literature; (iii) instead of discarding the oldest \textit{non-conformity scores}, discard according to a a better criteria (e.g., discard those that are less likely to belong the current data distribution); (iv) utilizing asymmetric and heteroscedasticity aware \textit{non-conformity measures} \citep{85}; and (v) employing ACI in conjunction with \texttt{EnbCQR} or \texttt{EnbPI}.

To conclude, \texttt{AEnbMIMOCQR} is a step further on extending CP for multi-step ahead forecasting in time series and may certainly encounter several applications in real-world sectors including, but not limited to, energy, finance, and retail.

\section*{CRediT authorship contribution statement}
\textbf{Martim Sousa:} Writing - original draft, Writing - review \& editing, Problem conceptualization, Methodology, Algorithm development, Investigation, Programming, Data analysis, Data curation. \textbf{Ana Tomé:} Writing - review \& editing, Supervision, Validation. \textbf{José Moreira}: Writing - review \& editing, Supervision, Validation, Funding acquisition.

\section*{Declaration of Competing Interest}
The authors declare that they have no known competing financial interests or personal relationships that could have appeared to influence the work reported in this paper.

\section*{Acknowledgments}
This work has been supported by COMPETE: POCI-01-0247-FEDER-039719 and FCT - Fundação para a Ciência e Tecnologia within the Project Scope: UIDB/00127/2020.

\appendix
\section{Proofs}
\label{proofs}

\begin{definition}[Exchangeability]
\label{exchdef}
A sequence of random variables $Z_1,Z_2,...,Z_n \in \mathcal{Z}$ are exchangeable if and only if for any permutation $\pi: \{1,2,...,n\} \rightarrow \{1,2,...,n\}$, we have
\begin{equation}
    \mathbb{P}\{Z_1,Z_2,...,Z_n\}= \mathbb{P}\{Z_{\pi(1)},Z_{\pi(2)},...,Z_{\pi(n)}\}
\end{equation}
\end{definition}

\begin{lemma}
\label{lemmabeforetheorem}
Let $Z_1,...,Z_n \in \mathcal{Z}$ be exchangeable random variables with no ties almost surely, then their ranks are uniformly distributed on \{1,...,n\}.
\begin{proof}
Let $R_i=\text{Rank}(Z_i)$. Since $Z_1,\ldots,Z_n$ are exchangeable, there are $n!$ equally probable permutations of the variables. Furthermore, because $Z_1,\ldots,Z_n$ have no ties almost surely, we have $\mathbb{P}(R_i\neq R_j)=1$ for all $i\neq j\in \{1,\ldots,n\}$.

For any $i\in\{1,...,n\}$, if we fix $R_i$ on rank $j\in \{1,...,n\}$, then there are (n-1)! possible permutations among the other n-1 random variables. 
    Hence, $\mathbb{P}(R_i=j)=\frac{(n-1)!}{n!}=\frac{1}{n}$ and thus
\begin{equation*}
    \mathbb{P}(R_i=1)=\mathbb{P}(R_i=2)=...=\mathbb{P}(R_i=n)=\frac{1}{n}, \quad \forall i \in\{1,...,n\}.
\end{equation*}
This shows that the ranks of the variables are uniformly distributed on ${1,\ldots,n}$.
\end{proof}
\end{lemma}

\begin{theorem}[Marginal coverage guarantee]
\label{theorem1}
Let $(\bm{X_1},Y_1),...,(\bm{X_{n}},Y_{n})$ be exchangeable random vectors with no ties almost surely drawn from a distribution P, additionally if for a new pair $(\bm{X_{n+1}},Y_{n+1})$, $(\bm{X_1},Y_1),...,(\bm{X_{n+1}},Y_{n+1})$ are still exchangeable, then by constructing $C_{1-\alpha}(\bm{X_{n+1}})$ using CP, the following inequality holds for any \textit{non-conformity score function} $s:\mathcal{X} \times \mathcal{Y} \rightarrow \mathcal{A} \subseteq \mathbb{R}$ and any $\alpha \in (\frac{1}{n+1},1)$
\begin{equation}
    1-\alpha\leq \mathbb{P}\{Y_{n+1} \in C_{1-\alpha}(\bm{X_{n+1}})\} \le 1-\alpha+\frac{1}{n+1}.
\end{equation}

\begin{proof}
Since $\bm{Z_1}=(\bm{X_1},Y_1),...,\bm{Z_n}=(\bm{X_n},Y_n)$ are exchangeable random vectors with no ties almost surely, the corresponding \textit{non-conformity scores} $\bm{\epsilon}= \{(\epsilon_i:=s(\bm{Z_i}))\}_{i=1}^{n} $ are also exchangeable (by Theorem 3 of \citep{54}) with no ties almost surely.
Since the non-conformity scores $\epsilon_{1},...,\epsilon_{n}, \epsilon_{n+1}$ are exchangeable, their ranks are uniformly distributed by Lemma (\ref{lemmabeforetheorem}). Therefore, $\text{Rank}(\epsilon_{n+1}) \sim \mathcal{U}\{1,n+1\}$. That is,

\begin{equation*}
    \mathbb{P}\{\epsilon_{n+1} \leq \epsilon_{(k)} \}=  \mathbb{P}\{\text{Rank}(\epsilon_{n+1}) \leq k \}=\frac{k}{n+1}, \quad  \forall k \in\{1,...,n+1\}.
\end{equation*}
By definition, since $\hat{q}=\text{Quantile}(\bm{\epsilon} ; \frac{\lceil (n+1)(1-\alpha)\rceil}{n})=\epsilon_{(\lceil(n+1)(1-\alpha) \rceil)}$, follows that
\begin{align*}
    &\mathbb{P}\{Y_{n+1} \in C_{1-\alpha}(\bm{X_{n+1})}\}=
    \mathbb{P}\{y \in \mathcal{Y}: s(\bm{X_{n+1}},y) \leq \hat{q}\}=\mathbb{P}\{\epsilon_{n+1}\leq \epsilon_{(\lceil(n+1)(1-\alpha) \rceil)}\}=\frac{\lceil(n+1)(1-\alpha) \rceil}{n+1}.
\end{align*}
It is easy to note that $1-\alpha \leq \frac{\lceil(n+1)(1-\alpha) \rceil}{n+1} \leq \frac{(n+1)(1-\alpha)+1}{n+1} = 1-\alpha+\frac{1}{n+1}$.
\end{proof}
\end{theorem}
\section{Algorithms}
\label{algos}

\begin{algorithm}[H]
\caption{\texttt{EnbPI} algorithm adapted to recursive $H$-step ahead PIs. \citep{12}}.
\label{EnbPIalg}
\begin{algorithmic}[1]
\Require Training data $\{(\bm{x_i},y_i)\}_{i=1}^n$, regression algorithm $\mathcal{A}$, miscoverage rate $\alpha$, aggregation function $\phi$, number of bootstrap models $B$, the forecast horizon $H$, and test data $\{(\bm{x_t},y_t)\}_{t=n+1}^{n+n_{\text{test}}}$, with $y_{t}$ revealed only after $H$ timesteps.
\Ensure PIs: $\{\hat{C}_{1-\alpha}^{(t)}(\bm{x_t})\}_{t=n+1}^{n + n_{\text{test}}}$
\For{b=1,...,B}
\State Sample an index set $S_b=(i_1,...,i_T)$ from indices (1,...,n)
\State Train $\hat{f}^{b} \gets \mathcal{A}(\{\bm{x_i}, y_i)\;|\;i \in S_b\})$
\EndFor
\State  $\bm{\epsilon} \gets\{\}$
\For{$i \gets 1,...,n$}
\State $\hat{y}_{i} \gets \phi(\{\hat{f}^{b}(\bm{x_i})\; | \; i \not\in S_b \})$
\State $ \epsilon_{i}^{\phi} \gets |\hat{y}_i-y_i|$
\State $\bm{\epsilon} \gets \bm{\epsilon} \cup \{\ \epsilon_{i}^{\phi}\}$
\EndFor
\State $\hat{q} \gets \text{Quantile} \left( \bm{\epsilon}; 1-\alpha\right)$ 
\For{$t \gets n+1,...,n+n_{\text{test}}$}
\State Compute $\hat{\bm{x_t}}$ using the recursive strategy formula (see (\ref{recstrategy})) for $h=t-T\mod H$
\State $\hat{y}_{t}\gets \phi( \{ \hat{f}^{b}(\bm{x_t})\}_{b=1}^B )$
\State Return $\hat{C}_{1-\alpha}^{(t)}(\bm{x_t}) \gets [\hat{y}_{t}-\hat{q}, \hat{y}_{t} +\hat{q}]$ 
\If{$t-T \equiv 0 \mod H$}
\For{$j \gets t-H,...,t-1$}
\State $\epsilon_{j}^{\phi}\gets|\hat{y}_{j}-y_{j}|$
\State $\bm{\epsilon} \gets (\bm{\epsilon}-\{\epsilon_1^{\phi}\}) \cup \{\epsilon_{j}^{\phi}\}$ and reset index of $\bm{\epsilon}$
\EndFor
\State  Update $\hat{q} \gets \text{Quantile} \left( \bm{\epsilon}; 1-\alpha\right)$ 
\EndIf
\EndFor
\end{algorithmic}
\end{algorithm}

\begin{algorithm}[H]
\caption{\texttt{EnbCQR} algorithm adapted to recursive $H$-step ahead PIs. \citep{7}}.
\label{EnbCQRalg}
\begin{algorithmic}[1]
\Require Training data $\{(\bm{x_i},y_i)\}_{i=1}^n$, regression single-output QR algorithm $\mathcal{A}_{\tau}$, miscoverage rate $\alpha$, aggregation function $\phi$, number of bootstrap models $B$, the forecast horizon $H$, and test data $\{(\bm{x_t},y_t)\}_{t=n+1}^{n+n_{\text{test}}}$, with $y_{t}$ revealed only after $H$ timesteps.
\Ensure PIs: $\{\hat{C}_{1-\alpha}^{(t)}(\bm{x_t})\}_{t=n+1}^{n + n_{\text{test}}}$
\For{b=1,...,B}
\State Sample an index set $S_b=(i_1,...,i_T)$ from indices (1,...,n)
\State Train $\hat{f}_{\alpha/2}^{b} \gets \mathcal{A}_{\alpha/2}(\{\bm{x_i}, y_i)\;|\;i \in S_b\})$
\State Train $\hat{f}_{0.5}^{b} \gets \mathcal{A}_{0.5}(\{\bm{x_i}, y_i)\;|\;i \in S_b\})$
\State Train $\hat{f}_{1-\alpha/2}^{b} \gets \mathcal{A}_{1-\alpha/2}(\{\bm{x_i}, y_i)\;|\;i \in S_b\})$
\EndFor
\State  $\bm{\epsilon} \gets\{\}$
\For{$i \gets 1,...,n$}
\State $\hat{y}_{i}^{L} \gets \phi(\{\hat{f}^{b}_{\alpha/2}(\bm{x_i})\; | \; i \not\in S_b \})$
\State $\hat{y}_{i}^{U} \gets \phi(\{\hat{f}^{b}_{1-\alpha/2}(\bm{x_i})\; | \; i \not\in S_b \})$
\State $ \epsilon_{i}^{\phi} \gets \max\{\hat{y}_{i}^{L}-y_i, y_i - \hat{y}_i^{U}\}$
\State $\bm{\epsilon} \gets \bm{\epsilon} \cup \{\ \epsilon_{i}^{\phi}\}$
\EndFor
\State $\hat{q} \gets \text{Quantile} \left( \bm{\epsilon}; 1-\alpha\right)$ 
\For{$t \gets n+1,...,n+n_{\text{test}}$}
\State Compute $\hat{\bm{x_t}}$ using $\{\hat{f}^{b}_{0.5}\}_{b=1}^B$ and the recursive strategy formula (see (\ref{recstrategy})) for $h=t-T\mod H$
\State $\hat{y}_{t}^{L}\gets \phi( \{ \hat{f}^{b}_{\alpha/2}(\hat{\bm{x_t}})\}_{b=1}^B )$
\State $\hat{y}_{t}^{U}\gets \phi( \{ \hat{f}^{b}_{1-\alpha/2}(\hat{\bm{x_t}})\}_{b=1}^B )$
\State Return $\hat{C}_{1-\alpha}^{(t)}(\bm{x_t}) \gets [\hat{y}_{t}^{L}-\hat{q}, \hat{y}_{t}^{U} +\hat{q}]$ 
\If{$t-T \equiv 0 \mod H$}
\For{$j \gets t-H,...,t-1$}
\State $ \epsilon_{j}^{\phi} \gets \max\{\hat{y}_{i}^{L}-y_i, y_i - \hat{y}_i^{U}\}$
\State $\bm{\epsilon} \gets (\bm{\epsilon}-\{\epsilon_1^{\phi}\}) \cup \{\epsilon_{j}^{\phi}\}$ and reset index of $\bm{\epsilon}$
\EndFor
\State  Update $\hat{q} \gets \text{Quantile} \left( \bm{\epsilon}; 1-\alpha\right)$ 
\EndIf
\EndFor
\end{algorithmic}
\end{algorithm}

\begin{algorithm}[H]
\caption{\texttt{MIMOCQR} algorithm to obtain $H$-step ahead PIs}.
\label{MIMOCQRalg}
\begin{algorithmic}[1]
\Require A training set $\{(\bm{x_i},\bm{y_i})\}_{i=1}^{n}$, a calibration set $\{(\bm{x_i},\bm{y_i})\}_{i=n+1}^{n+n_{\text{cal}}}$,  miscoverage rate $\alpha$, a multi-output QR algorithm $\mathcal{A}_{\tau}$, the forecast horizon $H$, and a test set $\{(\bm{x_t}, \bm{y_t})\}_{t=n+n_{\text{cal}}+1}^{n+n_{\text{cal}}+n_{\text{test}}}$ with $\bm{y_{t}}$ revealed only at timestep $t+H$.

\Ensure PIs $\{\hat{C}_{1-\alpha}^{(t)}(\bm{x_t})\}_{t=n+1}^{n+n_{\text{test}}}$

\State Train $\hat{F}_{\alpha/2} \gets \mathcal{A}_{\alpha/2}(\{(\bm{x_i},\bm{y_i})\;|\;i \in \{1,...,n\}\})$
\State Train $\hat{F}_{1-\alpha/2} \gets \mathcal{A}_{1-\alpha/2}(\{(\bm{x_i},\bm{y_i})\;|\;i \in \{1,...,n\}\})$
\State $\bm{\epsilon}_h \gets \{\}, \quad \forall h \in\{1,...,H\}$

\For{$i \gets n+1,...,n+n_{\text{cal}}$}
\State $[\hat{y}_{i,1}^{L},...,\hat{y}_{i,H}^{L}] \gets \hat{F}_{\alpha/2}(\bm{x_i})$
\State $[\hat{y}_{i,1}^{U},...,\hat{y}_{i,H}^{U}] \gets \hat{F}_{1-\alpha/2}(\bm{x_i})$
\State $ \epsilon_{i,h}^{\phi} \gets \max{\{\hat{y}_{i,h}^{L}-y_{i,h},y_{i,h}-\hat{y}_{i,h}^{U}\}}, \quad \forall h \in\{1,...,H\}$
\State $\bm{\epsilon}_h \gets \bm{\epsilon}_h \cup \{\epsilon_{i,h}^{\phi}\}, \quad \forall h \in\{1,...,H\}$
\EndFor
\State $\hat{q}^{(h)} \gets \text{Quantile}(\bm{\epsilon_h};1-\alpha), \quad \forall h \in\{1,...,H\}$
\For{$t\gets n+n_{\text{cal}}+1,n+n_{\text{cal}}+H+1,..., n+n_{\text{cal}}+n_{\text{test}}$}
\State $[\hat{y}_{t,1}^{L},...,\hat{y}_{t,H}^{L}] \gets \hat{F}_{\alpha/2}(\bm{x_t})$
\State $[\hat{y}_{t,1}^{U},...,\hat{y}_{t,H}^{U}] \gets \hat{F}_{1-\alpha/2}(\bm{x_t})$
\State Return $\hat{C}_{1-\alpha}^{(t)}(\bm{x_t}) \gets [\hat{y}_{t,h}^{L}-\hat{q}^{(h)},\hat{y}_{t,h}^{U}+\hat{q}^{(h)}], \quad \forall h \in\{1,...,H\}$
\EndFor

\end{algorithmic}
\end{algorithm}

\bibliography{refs.bib}

\end{document}